\documentclass{article}

\usepackage{microtype}
\usepackage[table]{xcolor}
\usepackage{graphicx}
\usepackage{subcaption}
\usepackage{cuted}
\usepackage{capt-of}
\usepackage{booktabs} 

\usepackage{hyperref}

\usepackage{algorithm}
\usepackage{algpseudocode} 

\usepackage[preprint]{icml2026}

\usepackage{amsmath}
\usepackage{amssymb}
\usepackage{mathtools}
\usepackage{amsthm}
\usepackage{multirow}

\usepackage[capitalize,noabbrev]{cleveref}

\theoremstyle{plain}

\theoremstyle{definition}

\theoremstyle{remark}

\newcommand{\cut}[1]{}

\DeclareMathOperator*{\argmax}{arg\,max}

\usepackage[textsize=tiny]{todonotes}

\icmltitlerunning{Tuning Out-of-Distribution (OOD) Detectors Without Given OOD Data}

\begin{document}

\twocolumn[
  \icmltitle{Tuning Out-of-Distribution (OOD) Detectors Without Given OOD Data}



  \icmlsetsymbol{equal}{*}

  \begin{icmlauthorlist}
    \icmlauthor{Sudeepta Mondal}{rtx}
    \icmlauthor{Xinyi Mary Xie}{yale}
    \icmlauthor{Ruxiao Duan}{yale}
    \icmlauthor{Alex Wong}{yale}
    \icmlauthor{Ganesh Sundaramoorthi}{rtx}
  \end{icmlauthorlist}

  \icmlaffiliation{rtx}{RTX}
  \icmlaffiliation{yale}{Yale University}

\icmlcorrespondingauthor{Ganesh Sundaramoorthi}{ganesh.sundaramoorthi@rtx.com}

  \icmlkeywords{Machine Learning, ICML}

  \vskip 0.3in
]



\printAffiliationsAndNotice{}  


\begin{abstract}
    Existing out-of-distribution (OOD) detectors are often tuned by a separate dataset deemed OOD with respect to the training distribution of a neural network (NN). OOD detectors process the activations of NN layers and score the output, where parameters of the detectors are determined by fitting to an in-distribution (training) set and the aforementioned dataset chosen adhocly. At detector training time, this adhoc dataset may not be available or difficult to obtain, and even when it's available, it may not be representative of actual OOD data, which is often ''unknown unknowns." Current benchmarks may specify some left-out set from test OOD sets. We show that there can be significant variance in performance of detectors based on the adhoc dataset chosen in current literature, and thus even if such a dataset can be collected, the performance of the detector may be highly dependent on the choice. In this paper, we introduce and formalize the often neglected problem of tuning OOD detectors without a given ``OOD'' dataset.  To this end, we present strong baselines as an attempt to approach this problem. Furthermore, we propose a new generic approach to OOD detector tuning that does not require any extra data other than those used to train the NN. We show that our approach improves over baseline methods consistently across higher-parameter OOD detector families, while being comparable across lower-parameter families.
    
  \end{abstract}

\section{Introduction}

\begin{figure}
    \centering
    \includegraphics[width=0.49\linewidth]{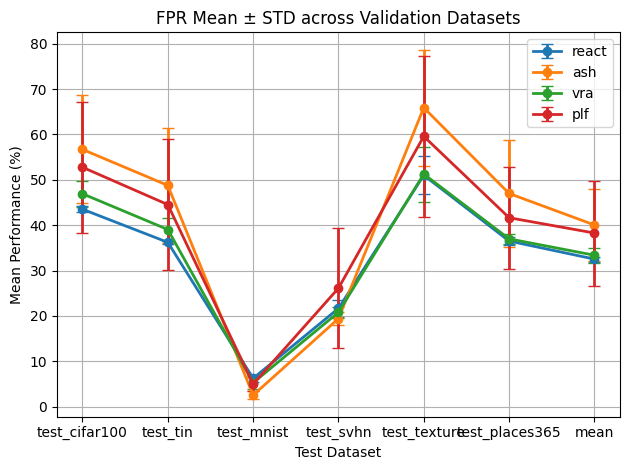}
    \includegraphics[width=0.49\linewidth]{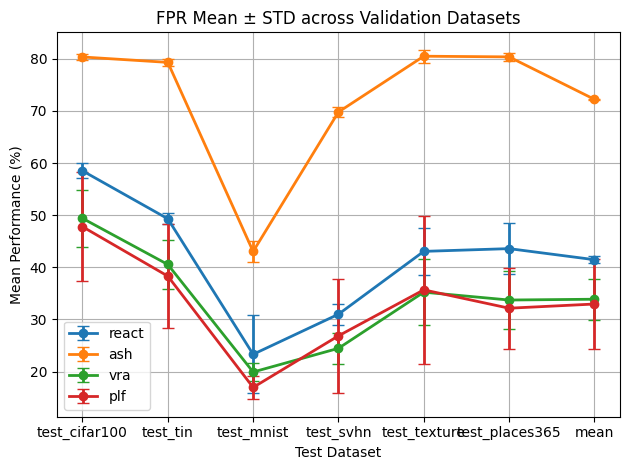}
    \caption{Performance of OOD detectors varies with tuning set. Several state-of-the-art OOD detectors parameters are tuned using different predefined tuning datasets deemed OOD in the OpenOOD benchmark. The mean and standard deviation of the detectors' FPR95 (y-axis) over tuning sets for each test OOD datasets (x-axis) is shown. Depending on the tuning dataset selected, there can be significant variability in performance of the same detectors (vertical bars), so much so that the rankings of detectors can change. Results on two different networks (DenseNet101 - left and ResNet-18 - right) are shown. Our approach avoids the need for given OOD data, which may be difficult to obtain in practice, and variance associated with the choice, and still has comparable performance to given test OOD sets as OOD tuning data in state-of-the-art benchmarks (see Table~\ref{tab:table_val_set_tuning_comparison}).
    }
    \label{fig:sensitivity_OOD_val_datasets}
\end{figure}

\comment{
\textbf{Story}
\begin{itemize}
    \item OOD detection is significant problem in using AI in practical applications
    \item Much of the focus has been on designing OOD detectors, especially shaping functions
    \item An important issue in using these methods in practice, often swept under the rug in the literature, is tuning parameters of the detector
    \item there is no systematic or agreed upon approach; existing benchmarks specify some held out OOD data from a OOD test set for parameter tuning / validation, but in practice it leaves the problem to the practioner, as 
    \begin{enumerate}
    \item OOD data may not generally be available before operation at deisgn time (and may be specific to the network as what is OOD depends not only on the data but also the network)
    \item may have significant cost to obtain it, depending on the application
    \item no guarantee that the validation data chosen sampled sufficiently the OOD space, and this choice may lead to variability of performance.  Figure~\ref{fig:sensitivity_OOD_val_datasets} shows that in the OpenOOD benchmark, the sensitivity of OOD detector performance to OOD val choice, in fact the best performing methods change depending on the validation set chosen, illustrating the difficulty of choosing validation data
    \end{enumerate}
    \item Therefore, a systematic and reliable approach to tuning OOD detectors is needed, and moreover avoiding the use of real OOD validation data (which can be both sensitive and may present practical problems in collection) 
\end{itemize}

Contributions:
\begin{enumerate}
    \item We introduce a novel approach to OOD parameter tuning that is fully tuned on existing in distribution data and hence avoids the need for collecting or generating OOD validation data.
    \item We show through extensive benchmarking on several datasets and model architectures that our approach meets or exceeds the performance of OOD detectors trained on any chosen real OOD validation data, without requiring any OOD validation data.
    \item The prior two facts demonstrate our approach is the first reliable OOD detector parameter tuning method that requires only in-distribution training data (and not OOD validation data).
\end{enumerate}
}

When machine learning models encounter inputs that differ in distribution from their training distribution, their predictions can become unreliable or unsafe, making accurate \emph{OOD detection} crucial, especially in safety-critical systems \cite{yang2022generalized}. As a result, a large body of research has focused on designing increasingly sophisticated OOD detectors, including score-based methods \cite{liu2021energy}, density models \cite{Zong2018DeepAG,he2015deepresiduallearningimage}, and learned shaping functions \cite{sun2021reactoutofdistributiondetectionrectified, djurisic2022ash,mondal2025variationalinformationtheoreticapproach}.

Despite this progress, an important practical challenge has received surprisingly little systematic attention: how to tune the parameters of OOD detectors. In nearly all modern OOD methods, performance depends critically on one or more parameters that control, for example, thresholds or shaping functions. Existing benchmarks and experimental protocols typically assume access to a held-out dataset deemed OOD with respect to the training data of a neural network for a task of interest, which we refer to as a ``task NN''. This OOD tuning dataset is used to determine the parameters of a OOD detector before deployment onto the task NN for evaluation. Because it is difficult to foresee the data to be encountered at test time, the choice of this tuning dataset is typically adhoc. While this is convenient for benchmarking, it masks a serious gap between experimental practice and real-world deployment.

In practice, a suitable tuning dataset is rarely available. What constitutes ``out-of-distribution'' depends not only on the data domain but also on the specific trained model, making it difficult to predefine representative OOD samples. Moreover, collecting or generating realistic OOD data can be costly, sensitive, or operationally infeasible in many applications. Even when some OOD samples are available, there is no guarantee that they sufficiently cover the space of possible OOD inputs, leading to substantial variability in the subsequent detection outcome. Figure~\ref{fig:sensitivity_OOD_val_datasets} illustrates this issue on the OpenOOD benchmark \cite{zhang2024openoodv15enhancedbenchmark}, where the relative ranking of state-of-the-art OOD detectors changes depending on which OOD tuning set is used. 

In this work, we investigate the problem of tuning OOD detectors without being given a dataset. To the best of our knowledge, the only attempt of tuning OOD detectors without an adhoc tuning set is through the use of Gaussian noise images \cite{kirichenko2020normalizing}. While not in the OOD detection literature, we hypothesize that additive adversarial perturbations \cite{goodfellow2015explaining} can also serve as a strong baseline. We begin by benchmarking these baseline methods as a means to generate synthetic or simulated tuning sets for determining the parameters of OOD detectors. We note that these choices, too, are largely adhoc and found that they are only effective as tuning sets in limited settings, e.g., Gaussian noise images were specifically used to tune OOD detectors on ImageNet \cite{sun2021reactoutofdistributiondetectionrectified, xu2023vravariationalrectifiedactivation}. Hence, we address this gap by introducing a new paradigm for OOD detector tuning that eliminates the need for a tuning dataset altogether. Our key insight is that the structure of the in-distribution (used to train the task NN) itself contains sufficient information to calibrate OOD detectors, if it is exploited in the right way. This leads to a principled and fully self-contained tuning procedure that depends only on the data already available during the training of the neural network on which the OOD detector will be deployed.

\paragraph{Contributions.}
\begin{enumerate}
    \item We introduce and formalize the problem of the determing the parameters of OOD detectors without access to a separate tuning dataset.
    \item We systematically evaluate baselines for this problem, and find no consistent best performer across standard benchmark datasets for multiple NN architectures.
    \item We present a novel OOD parameter tuning method that uses only in-distribution training data of the task NN and does not require any adhoc tuning dataset.
    \item Amongst the evaluated methods, we show that our approach consistently outperforms baselines in higher-parameter OOD detector families, while all methods are comparable in lower-parameter families. 
\end{enumerate}

\subsection{Related Work}
We briefly review related work; the reader is referred to \cite{yang2022generalized} for a survey. Our focus is on post-hoc inference methods of OOD detection~\cite{zhang2024openoodv15enhancedbenchmark}, which are applied to pre-trained models without additional training.  These methods construct scoring functions that aim to distinguish in-distribution (ID) from OOD inputs using properties of the model’s outputs or intermediate representations. Prominent examples include confidence-based scores \cite{hendrycks2018,zhang_decoupling,odin_liang}, energy-based metrics \cite{liu2021energy,wu2023energy,elflein2021out}, and distance-based measures in feature space \cite{lee2018simple,sun2022knn}.

Early work such as MSP \cite{hendrycks2018} used the maximum softmax probability as a confidence score, while ODIN \cite{odin_liang} improved upon this by introducing temperature scaling and input perturbations. Likelihood-based approaches using deep generative models have also been explored, although raw likelihoods are known to be unreliable for OOD detection \cite{kirichenko2020normalizing}, motivating alternatives such as likelihood ratios \cite{ren2019likelihood}. Distance-based methods measure deviation from class-conditional feature distributions, for example using Mahalanobis distance \cite{lee2018simple} or non-parametric nearest-neighbor schemes \cite{sun2022knn}. Energy-based scoring \cite{liu2021energy} provides a unified alternative to softmax confidence by interpreting logits through the Helmholtz free energy, and has since been adopted by several feature-based OOD detectors.

{\bf Feature-shaping approaches to OOD detection}: Several OOD detection methods operate by transforming intermediate network features prior to scoring \cite{sun2021reactoutofdistributiondetectionrectified, zhao2024towards}. ReAct \cite{sun2021reactoutofdistributiondetectionrectified} applies an element-wise clipping operation to the penultimate layer activations, motivated by the empirical observation that OOD inputs often induce unusually large activation values. ASH~\cite{djurisic2022ash} performs feature shaping through sparsification, setting small activations to zero while optionally rescaling larger ones. DICE \cite{sun2022dice} follows a similar sparsification principle. Unlike purely element-wise transformations, ASH additionally applies a vector-level normalization step before computing the detection score. In contrast, VRA \cite{xu2023vravariationalrectifiedactivation} and related work \cite{zhanglearning} construct element-wise shaping functions via an explicit optimization procedure.

Across the broad literature, we find that OOD detectors often rely on adhoc tuning dataset(s) for determining their parameters. These OOD tuning datasets are also typically paired with a specific target dataset, e.g., TIN, MNIST and SVHN for CIFAR10 and CIFAR100, and iNaturalist, OpenImage-O and NINCO with ImageNet-200 and ImageNet-1K. Attempts to depart from the reliance of these tuning datasets have been limited, and are also adhoc and specific to the target dataset, e.g., Gaussian noise images for ImageNet-1K \cite{sun2021reactoutofdistributiondetectionrectified, xu2023vravariationalrectifiedactivation}; we treat these as baselines. Our work aims to present the first general framework for OOD tuning detectors without the need for any predefined OOD tuning dataset. Our method is broadly applicable across task NN training datasets and OOD detectors.

\section{Our Method}

In this section, we present our method for tuning OOD detector parameters without being given a tuning dataset. We will develop and illustrate our approach for the classification problem. We denote the in-distribution training dataset $\mathcal D^{t} = \{\mathcal C_1^t, \ldots, \mathcal C_n^t\}$ where $n$ is the number of categories and $\mathcal C_i^t$ is the set of data for category $i$. 

The crux of the method lies in observation that out-of-distribution (OOD) data lies any where outside of the training (ID) distribution within some latent feature space. Given a stationary training dataset, data samples held out from the training process naturally lie ``close'' to the ID distribution. Hence, these held-out data can be seen as hard examples. By tuning the parameters of a function to discriminate between these hard (OOD) examples and those of the ID distribution, we hypothesize that the OOD detector can robustly identify both ``near'' and ``far'' OOD examples in novel datasets. To this end, our method aims to simulate ID and OOD examples from the training dataset of the task NN, which allows us to bypass the need for a separate tuning dataset to determine the parameters of the OOD detector.

{\bf Simulated ID/OOD Datasets from ID Training Data}: Given a neural network architecture $f_{\theta}$ (task NN), where $\theta$ represents the parameters of the network, we train $N$ variants of $f_{\theta}$ for the original task (classification) as follows. Choose $M :=  M_{ood,sim}$ categories at random from the $n$ categories; we refer to these as the \emph{simulated OOD categories}; the data from these categories is denoted $\mathcal D_{ood,sim}^M$. We refer to the remaining $M_{id,sim} = n - M_{ood,sim}$ categories as the \emph{simulated ID categories}, the data is denoted $\mathcal D_{id,sim}^M$. We train the network $f_{\theta_1}^M$ for the task (classification) on $\mathcal D_{id,sim}^M$.  This process is repeated $N$ times over different random choices of simulated ID/OOD categories to create $f_{\theta_1}^M, \ldots, f_{\theta_N}^M$ variants of the trained network; we also denote $\mathcal D_{id,sim,i}^M$ ($\mathcal D_{ood,sim,i}^M$) the corresponding simulated ID (OOD) data used to train $f_{\theta_i}^M$. The networks $f_{\theta_i}^M$ will all share similarity to the network trained on all data if $M_{ood,sim}$ is small.  The dataset $\mathcal D_{ood,sim,i}^M$ will now be OOD to the network $f_{\theta_i}^M$ as this data is excluded from training the network. Furthermore, we conjecture that that data will be near ID since it's from the same dataset underlying dataset collected with the same sampling process, but OOD since the network has not ''seen" it or data from that category. We conjecture that tuning an OOD detector on such a simulated dataset(s) will learn the true OOD/ID boundary.

{\bf Generating Tuning Sets for OOD Detector Training}: We now specify the procedure to tune the OOD detector. Given an OOD detector $d_{\phi}$ with parameters $\phi$ that operates on a task neural network, $f_{\theta}$, e.g., shaping of the penultimate layer features. We now sample $S$ simulated ID/OOD tuning sets $\mathcal D_{id,sim,i,j}^M$ and $\mathcal D_{ood,sim,i,j}^M$ from the simulated ID data $\mathcal D_{id,sim,i}^M$ and simulated OOD data $\mathcal D_{ood, sim, i}^M$. These sets and the task networks $f_{\theta_i}^M$ are used to train the OOD detector $d_{\phi}$. To train $d_{\phi}$ we use $\mathcal D_{id,sim,i,j}^M$ as samples of ID data and $\mathcal D_{ood,sim,i,j}^M$ as samples of OOD data \footnote{Note in our choices $M_{ood,sim}$ is much less than $M_{id,sim}$ and thus there is an imbalance of simulated ID and OOD tuning data. To alleviate this, we use the OOD tuning and training data (which the network has not seen) to create balanced simulated datasets.}. The algorithm for simulation of the simulated ID/OOD data and the simulated tuning data is shown in Algorithm~\ref{alg:sim_datasets}.

\begin{algorithm}
\caption{Simulated ID/OOD Training and OOD Detector Validation Data from ID Classification Training Data}
\label{alg:sim_datasets}
\begin{algorithmic}[1]
\Require The network training set $\mathcal D^{t} = \{\mathcal C_1^t, \ldots, \mathcal C_n^t\}$
\Ensure Trained networks $\{f_{\theta_1}^M, \ldots, f_{\theta_N}^M\}$,
        simulated validation datasets $\mathcal D_{\text{id,sim},i,j}^M$
        and $\mathcal D_{\text{ood,sim},i,j}^M$, simulated ID/OOD datasets $\mathcal D_{\text{id,sim},i}^M$ and
        $\mathcal D_{\text{ood,sim},i}^M$
\For{each $M$}
    \For{$i = 1,\ldots,N$}
        \State Randomly sample $M$ simulated OOD categories to form
         $\mathcal{D}_{\text{ood,sim},i}^M$
        \State Define remaining classes as simulated ID dataset
        $\mathcal{D}_{\text{id,sim},i}^M$
        \State Train classification network $f_{\theta_i}^M$ on $\mathcal{D}_{\text{id,sim},i}^M$
        \For{$j = 1,\ldots,S$}
            \State Sample simulated validation ID dataset
            \Statex \hspace{1em} $\mathcal{D}_{\text{id,sim},i,j}^M \sim \mathcal{D}_{\text{id,sim},i}^M$
            \State Sample simulated validation OOD dataset
            \Statex \hspace{1em} $\mathcal{D}_{\text{ood,sim},i,j}^M \sim \mathcal{D}_{\text{ood,sim},i}^M$
        \EndFor
    \EndFor
\EndFor
\end{algorithmic}
\end{algorithm}

{\bf Loss Function for OOD Detector Training}: As training on a single sample $i$ of simulated data and the network $f_{\theta_i}$ and a single sample $j$ of ID/OOD tuning set is prone to randomness of choice, we tune the detector to detect simulated OOD well on average across all the $N$ simulated datasets/networks and $S$ of the tuning sets, as follows. Let $L(d_{\phi}; f_{\theta_i}^M, \mathcal D_{id, sim,i,j}^M, \mathcal D_{ood, sim,i,j}^M)$ denote the loss function associated with the loss of $d_{\phi}$ (operating on $f_{\theta_i}^M$) detecting simulated OOD data from simulated ID/OOD data. This could be metrics such as negative AUROC or FPR95 or a combination, etc. We minimize the loss $\ell$ defined by
\begin{equation} \label{eq:loss}
    \ell (\phi|M) =
    \frac{1}{N}
    \sum_{i=1}^N
    \frac{1}{S}
    \sum_{j=1}^S
   L(d_{\phi}; f_{\theta_i}^M, \mathcal D_{id,sim,i,j}^M, \mathcal D_{ood,sim,i,j}^M ),
\end{equation}
where $i$ indexes over the sampled left out categories and $j$ indexes over the sampled simulated ID/OOD tuning sets (sampled from the held-in/out data respectively). Note the inner sum estimates an expectation over simulated tuning sets of a particular choice of ID/OOD simulated categories, and the outer sum estimates an expectation over simulated ID/OOD categories. These expectations reduce variance and dependence on a particular simulated dataset choice. We perform a parameter search to minimize this loss. We choose to use Bayesian optimization \cite{frazier2018tutorialbayesianoptimization} to illustrate the idea, but any other technique can be used. The algorithm for optimization of this loss function is given in Algorithm~\ref{alg:optimal_parameters}.

{\bf Choosing the Number of Simulated OOD Categories}: Note the dependence of $\ell$ on the number $M$ of simulated OOD categories chosen. This is a hyper-parameter in our optimization scheme. Intuitively leaving out too few categories as OOD will not learn enough about the possible OOD distribution, while leaving out too many would mean that $f_{\theta_i}^M$ differs too much from the network trained on all the true ID categories. We offer an approach to choose the hyper-parameter $M$, which we use in experiments. We optimize $\ell(\phi|M)$ in $\phi$ to obtain $\phi^{\ast}_{M}$ for each $M$. We then re-sample datasets $\mathcal D^M_{id,sim,i,j}$ and $\mathcal D^M_{ood,sim,i,j}$ from $\mathcal D^M_{id,sim,i}$ and $\mathcal D^M_{ood,sim,i}$, and re-compute $\ell(\phi^{\ast}_{M}|M)$. This estimates the loss on independently generated simulated datasets following the same distribution as the simulated ID/OOD dataset, hence it can be used to validate the hyper-parameter choice. Therefore, we choose $M^{\ast}$ as
\begin{equation}
    M^{\ast} = \argmax_{M} \hat \ell(\phi^{\ast}_{M}|M),
\end{equation}
where $\hat \ell$ indicates that \eqref{eq:loss} is re-computed on new sampled datasets, as discussed previously.  The algorithm for hyper-parameter selection is given in Algorithm~\ref{alg:optimal_hyperparameter}.

{\bf Final Algorithm to Produce Tuned OOD Detector}: Starting from the original task network training set $\mathcal D^t$ and the trained network $f_{\theta}$ on all of $\mathcal D^t$, the simulated ID/OOD datasets and validation sets are generated by Algorithm~\ref{alg:sim_datasets}, which also produces trained networks $f_{\theta_i}^M$. Candidate optimal OOD detectors' parameters $\phi^{\ast}_M$ are obtained using Algorithm~\ref{alg:optimal_parameters}. Then Algorithm~\ref{alg:optimal_hyperparameter} is used to find the optimal hyper-parameter $M^{\ast}$ (number of left-out sets). The final OOD detector to be used in practice is now $\phi^{\ast}_{M^{\ast}}$ in conjunction with the original network $f_{\theta}$. This is shown in Algorithm~\ref{alg:tuning_OOD_detector}.





\begin{algorithm}[t]
\caption{Optimal OOD Detector Parameters $\phi$}
\label{alg:optimal_parameters}
\begin{algorithmic}[1]
\Require Simulated tuning datasets $\mathcal D_{\text{id,sim},i,j}^M$ and $\mathcal D_{\text{ood,sim},i,j}^M$
\Ensure Optimal detector parameters $\phi^{\ast}_M$

\For{each $M$}
    \State $\phi^{\ast}_M \leftarrow \arg\max_{\phi} \, \ell(\phi \mid M)$
    \Statex \hfill (Bayesian Optimization)
\EndFor

\end{algorithmic}
\end{algorithm}

\begin{algorithm}[t]
\caption{Optimal Hyper-parameter $M$}
\label{alg:optimal_hyperparameter}
\begin{algorithmic}[1]
\Require Simulated ID/OOD datasets $\mathcal D_{\text{id,sim},i}^M$,
         $\mathcal D_{\text{ood,sim},i}^M$
\Ensure Optimal hyper-parameter $M^{\ast}$

\For{each $M$}
    \State Resample $\mathcal D_{\text{id,sim},i,j}^M$ and $\mathcal D_{\text{ood,sim},i,j}^M$ from $\mathcal D_{\text{id,sim},i}^M$ and $\mathcal D_{\text{ood,sim},i}^M$    
    \State Compute validation loss $\hat{\ell}(\phi^{\ast}_M \mid M)$
\EndFor

\State $M^{\ast} \leftarrow \arg\max_M \hat{\ell}(\phi^{\ast}_M \mid M)$
\end{algorithmic}
\end{algorithm}

\begin{algorithm}[t]
\caption{Tuning OOD Detector Without Given OOD Data}
\label{alg:tuning_OOD_detector}
\begin{algorithmic}[1]
\Require The network training set $\mathcal D^{t} = \{\mathcal C_1^t, \ldots, \mathcal C_n^t\}$, task network $f_{\theta}$, detector $d_{\phi}$
\Ensure Tuned parameters $\phi^{\ast}_{M^{\ast}}$
\State Determine simulated ID/OOD sets and tuning sets - Algorithm~\ref{alg:sim_datasets}
\State Determine candidate optimal parameters $\phi^{\ast}_M$ - Algorithm~\ref{alg:optimal_parameters}
\State Determine optimal parameter set $\phi^{\ast}_{M^{\ast}}$ - Algorithm~\ref{alg:optimal_hyperparameter}
\end{algorithmic}
\end{algorithm}

\section{Baseline OOD Detector Tuning Methods}

In the experiments in the next section, we compare our new OOD detector tuning method without given OOD data to other baselines that have been considered in OOD detection papers or related literature in the past.  We emphasize though that there has been no formal treatment or formulation in the literature of the OOD detector tuning problem that does not use held out OOD tuning data nor any systematic evaluation of these baselines.  One of our contributions in this paper is to formalize the problem, formally introduce these baselines, and evaluate them.

The first baseline that we consider is using ID validation data from the validation set of the network training set and to use \emph{Gaussian noise images} as OOD validation used in \cite{sun2021reactoutofdistributiondetectionrectified}. That is Gaussian noise images, which are created by sampling Gaussian noise iid at each pixel~\cite{kirchheim2022pytorch}. The OOD detector parameters are trained with this simulated OOD dataset.

The second baseline is \emph{adversarial perturbed ID data}. Adversarial perturbations are visually imperceptible additive signals optimized by gradient ascent of the task training loss function. When added to data, these perturbations shift the sample into low-density regions of the data manifold \cite{song2018pixeldefend,stutz2019disentangling}. Thus, adversarial perturbations simulate OOD examples as the resulting data point exhibit lower likelihood under the training distribution. In our experiments, the ID data is sampled from the tuning data of the network training dataset. The OOD data is now the result of adversarially perturbing the ID tuning data using fast gradient sign method \cite{goodfellow2015explaining} to create the OOD tuning data.  Notice both the baselines do not rely on "true" OOD tuning data (often part of OOD benchmarks) or chosen expertly on a per-application basis.

Note that both baselines approaches have a hyper-parameter, in the case of Gaussian noise, the hyper-parameter is the noise level $\sigma$. This is not discussed in the literature and a hand-picked choice is presumably chosen. Our results show significant performance variation for the level of noise chosen. Similarly, the adversarial perturbation approach also has a parameter, the norm of the perturbation to be added. This is also often chosen by hand.

So there is a significant practical issue left out on these baselines in the literature, i.e., that of choosing the hyper-parameter in generating the "simulated" OOD validation data. Simply picking a value is not representative of the performance of the method given the variance in test results. Moreover, in practical applications one rarely has access to a test set to maximize performance on the test set. Therefore, we propose now a principled hyper-parameter tuning approach, analogous to our hyper-parameter tuning method introduced in the last section. We generate the ID and OOD validation data as specified previously for several values of the hyper-parameter, which we call $h$ (Gaussian noise level or norm of perturbation). Then we define the loss
\begin{equation}
    \ell(\phi|h) = \frac 1 S \sum_{j=1}^S
    L(d_{\phi}; f_{\theta}, \mathcal{D}_{id,sim,j}^h, 
    \mathcal{D}_{ood,sim,j}^h),
\end{equation}
where we sample $S$ datasets $\mathcal{D}_{id,sim,j}^h$ and $\mathcal{D}_{ood,sim,j}^h$ from the ID data and the simulated OOD data (Gaussian noise or adversarially perturbed images). This loss is optimized via Bayesian optimization to find $\phi^{\ast}_h$. Similarly to hyper-parameter tuning in the previous section, we resample the simulated ID/OOD data and then optimize the new loss $\ell(\phi^{\ast}_h|h)$ over $h$.

Therefore, in the comparisons of the baselines with our new method in the next section, we use our novel hyper-parameter tuning approach to choose the hyper-parameter for all methods with the same principled approach, which offers a systematic approach to hyper-parameter tuning in a practical setting.

\section{Experiments}

\begin{table*}[t]
\centering
\resizebox{\textwidth}{!}{
\begin{tabular}{lcccccccccccccc}
\toprule
 & \multicolumn{7}{c}{CIFAR-10 (ResNet-18)} & \multicolumn{7}{c}{CIFAR-100 (ResNet-18)} \\
\cmidrule(lr){2-8}
\cmidrule(lr){9-15}
Method & cifar100 & tin & mnist & svhn & texture & places365 & Best (\#) & cifar10 & tin & mnist & svhn & texture & places365 & Best (\#) \\
\midrule
ReAct (Gauss.) & 85.80 ± 0.81 & 88.30 ± 0.49 & 93.20 ± 3.51 & 88.92 ± 3.66 & 88.92 ± 1.15 & 90.51 ± 0.57 & 0 & 69.80 ± 2.25 & 77.28 ± 2.19 & 66.34 ± 4.46 & \textbf{83.97 ± 6.05} & \textbf{84.95 ± 0.72} & 78.28 ± 2.60 & 2 \\
ReAct (Adv.) & 85.51 ± 1.02 & 88.06 ± 0.69 & 92.90 ± 3.55 & 88.14 ± 3.92 & 88.59 ± 1.43 & \textbf{90.67 ± 0.71} & 1 & 75.13 ± 2.74 & 80.87 ± 1.38 & 72.96 ± 5.39 & 83.25 ± 4.17 & 82.92 ± 3.79 & 80.20 ± 0.69 & 0 \\
ReAct (Ours) & \textbf{86.42 ± 0.82} & \textbf{88.64 ± 0.64} & \textbf{93.87 ± 2.44} & \textbf{91.02 ± 1.38} & \textbf{89.13 ± 1.36} & 89.87 ± 1.18 & 5 & \textbf{78.16 ± 0.05} & \textbf{82.88 ± 0.07} & \textbf{77.28 ± 1.94} & 83.68 ± 1.28 & 81.57 ± 0.35 & \textbf{80.49 ± 0.08} & 4 \\
\hline
ASH (Gauss.) & 74.19 ± 3.31 & 76.69 ± 4.16 & 87.78 ± 2.94 & 73.33 ± 5.45 & 75.33 ± 6.18 & 80.25 ± 2.18 & 0 & 78.42 ± 1.28 & \textbf{80.76 ± 0.83} & 78.71 ± 5.54 & 81.10 ± 1.60 & 77.36 ± 1.71 & 78.30 ± 1.41 & 1 \\
ASH (Adv.) & \textbf{77.03 ± 2.11} & \textbf{79.36 ± 1.19} & \textbf{89.08 ± 3.29} & \textbf{80.98 ± 3.03} & \textbf{79.55 ± 1.27} & 79.72 ± 1.28 & 5 & \textbf{79.17 ± 0.24} & 80.67 ± 0.68 & \textbf{83.07 ± 0.40} & 80.00 ± 2.42 & 75.05 ± 1.46 & 77.41 ± 0.74 & 2 \\
ASH (Ours) & 76.77 ± 3.12 & 79.20 ± 2.67 & 85.82 ± 3.99 & 76.28 ± 6.71 & 78.37 ± 4.84 & \textbf{82.36 ± 1.97} & 1 & 76.49 ± 0.22 & 80.29 ± 0.23 & 73.81 ± 1.84 & \textbf{84.38 ± 1.82} & \textbf{79.32 ± 0.47} & \textbf{79.80 ± 0.10} & 3 \\
\hline
KNN (Gauss.) & 89.55 ± 0.12 & 91.34 ± 0.24 & 93.97 ± 0.36 & 92.42 ± 0.30 & 92.91 ± 0.24 & 91.52 ± 0.23 & 0 & 76.24 ± 0.27 & 82.91 ± 0.18 & \textbf{83.00 ± 1.44} & 84.00 ± 1.34 & \textbf{84.21 ± 0.80} & 78.66 ± 0.49 & 2 \\
KNN (Adv.) & 89.55 ± 0.12 & 91.34 ± 0.24 & 93.97 ± 0.36 & 92.42 ± 0.30 & 92.91 ± 0.24 & 91.52 ± 0.23 & 0 & 76.24 ± 0.27 & 82.92 ± 0.18 & 83.00 ± 1.44 & 84.00 ± 1.34 & 84.20 ± 0.80 & 78.67 ± 0.49 & 0 \\
KNN (Ours) & \textbf{89.60 ± 0.13} & \textbf{91.39 ± 0.25} & \textbf{94.04 ± 0.36} & \textbf{92.48 ± 0.30} & \textbf{92.97 ± 0.24} & \textbf{91.58 ± 0.23} & 6 & \textbf{76.65 ± 0.27} & \textbf{83.16 ± 0.18} & 82.64 ± 1.47 & \textbf{84.03 ± 1.18} & 83.98 ± 0.81 & \textbf{79.11 ± 0.48} & 4 \\
\hline
PLF (Gauss.) & 82.47 ± 3.29 & 84.88 ± 3.26 & 89.78 ± 3.51 & 87.80 ± 4.15 & 87.61 ± 3.46 & 83.55 ± 2.24 & 0 & 65.44 ± 3.16 & 72.79 ± 3.60 & 51.75 ± 7.18 & 67.61 ± 11.72 & 74.47 ± 11.72 & 73.55 ± 4.91 & 0 \\
PLF (Adv.) & 84.70 ± 0.23 & 87.02 ± 0.17 & 90.69 ± 3.66 & 87.17 ± 4.67 & \textbf{89.17 ± 1.39} & 86.55 ± 3.20 & 1 & 77.99 ± 0.64 & 82.42 ± 0.31 & 76.61 ± 3.65 & 81.84 ± 2.94 & 80.94 ± 4.19 & \textbf{80.62 ± 1.66} & 1 \\
PLF (Ours) & \textbf{85.55 ± 0.80} & \textbf{87.95 ± 0.45} & \textbf{94.30 ± 2.38} & \textbf{91.59 ± 1.34} & 88.53 ± 0.65 & \textbf{88.11 ± 0.83} & 5 & \textbf{78.72 ± 0.10} & \textbf{83.13 ± 0.07} & \textbf{78.71 ± 1.70} & \textbf{83.18 ± 0.57} & \textbf{81.26 ± 0.54} & 80.58 ± 0.14 & 5 \\
\hline
VRA (Gauss.) & 71.82 ± 9.74 & 72.49 ± 11.63 & 78.23 ± 8.32 & 74.71 ± 8.09 & 77.96 ± 8.99 & 71.74 ± 16.20 & 0 & 61.08 ± 3.13 & 68.07 ± 3.29 & 57.16 ± 4.97 & 74.79 ± 9.37 & 80.59 ± 2.28 & 70.28 ± 2.38 & 0 \\
VRA (Adv.) & 85.68 ± 0.81 & 87.94 ± 0.83 & 92.50 ± 3.37 & 88.88 ± 3.36 & 89.46 ± 0.97 & \textbf{89.71 ± 1.75} & 1 & 75.07 ± 3.38 & 80.73 ± 2.17 & 73.30 ± 5.07 & 83.40 ± 3.38 & \textbf{83.42 ± 3.20} & 78.94 ± 0.91 & 1 \\
VRA (Ours) & \textbf{86.22 ± 0.62} & \textbf{88.57 ± 0.37} & \textbf{94.10 ± 2.46} & \textbf{91.49 ± 1.06} & \textbf{89.71 ± 0.78} & 88.98 ± 0.66 & 5 & \textbf{77.40 ± 0.24} & \textbf{82.71 ± 0.05} & \textbf{76.89 ± 2.43} & \textbf{84.05 ± 1.52} & 83.37 ± 0.17 & \textbf{80.15 ± 0.21} & 5 \\
\hline
\bottomrule
\end{tabular}
}
\caption{Benchmark results on CIFAR-10 and CIFAR-100 ID datasets using OpenOOD test OOD datasets~\cite{zhang2024openoodv15enhancedbenchmark}. We report AUROC (higher is better). Bold indicates the best-performing detector–tuning method for each ID model. Best (\#) denotes the number of test OOD datasets on which a method achieves the highest AUROC. Gauss denotes the Gaussian noise baseline, Adv denotes the adversarial noise baseline, and ours indicates our tuning method.}

\label{tab:cifar_results}
\end{table*}

\begin{table*}[t]
\centering
\resizebox{\textwidth}{!}{
\begin{tabular}{lcccccc|cccccc|cccccc}
\toprule
 & \multicolumn{6}{c|}{ImageNet-200 (ResNet-18)} 
 & \multicolumn{6}{c|}{ImageNet-1K (ResNet-50)} 
 & \multicolumn{6}{c}{ImageNet-1K (MobileNet-V2)} \\
\cmidrule(lr){2-7}
\cmidrule(lr){8-13}
\cmidrule(lr){14-19}

Method 
& SSB-hard & NINCO & iNaturalist & Textures & OpenImage-O & Best (\#)
& SSB-hard & NINCO & iNaturalist & Textures & OpenImage-O & Best (\#)
& SSB-hard & NINCO & iNaturalist & Textures & OpenImage-O & Best (\#) \\
\midrule

ReAct (Gauss.) 
& 77.85 ± 2.81 & 83.59 ± 2.13 & \textbf{94.03 ± 0.51} & \textbf{93.44 ± 0.15} & \textbf{90.88 ± 0.30} & 3
& 73.15 & \textbf{81.94} & \textbf{96.07} & \textbf{92.73} & \textbf{91.93} & 4
& 60.98 & 72.29 & 89.60 & 91.46 & 84.45 & 0 \\

ReAct (Adv.) 
& 79.25 ± 0.33 & 84.97 ± 0.28 & 92.85 ± 0.90 & 91.50 ± 0.86 & 88.93 ± 0.16 & 0
& 72.48 & 80.46 & 92.40 & 89.73 & 89.85 & 0
& \textbf{61.91} & 73.17 & \textbf{92.91} & \textbf{94.12} & \textbf{86.81} & 4 \\

ReAct (Ours) 
& \textbf{79.83 ± 0.02} & \textbf{85.17 ± 0.12} & 92.56 ± 0.50 & 90.87 ± 0.16 & 89.25 ± 0.25 & 2
& \textbf{73.18} & 81.69 & 94.88 & 91.72 & 91.33 & 1
& 61.69 & \textbf{73.51} & 92.16 & 93.69 & 86.56 & 1 \\

\hline

ASH (Gauss.) 
& 77.06 ± 0.43 & 81.46 ± 0.76 & 93.31 ± 1.02 & 94.61 ± 0.06 & 89.68 ± 0.85 & 1
& \textbf{73.61} & 82.06 & 95.62 & 95.03 & 91.71 & 1
& \textbf{61.24} & \textbf{75.31} & \textbf{92.89} & \textbf{96.66} & \textbf{88.69} & 5 \\

ASH (Adv.) 
& \textbf{79.69 ± 0.15} & \textbf{85.40 ± 0.57} & \textbf{95.06 ± 0.63} & \textbf{94.97 ± 0.06} & \textbf{91.93 ± 0.39} & 5
& 73.00 & \textbf{83.38} & \textbf{96.88} & \textbf{96.71} & \textbf{93.20} & 4
& 61.07 & 74.81 & 91.85 & 96.35 & 88.05 & 0 \\

ASH (Ours) 
& 79.38 ± 0.29 & 84.51 ± 0.43 & 94.52 ± 0.41 & 93.14 ± 0.17 & 90.42 ± 0.09 & 0
& 73.27 & 83.17 & 96.60 & 96.29 & 92.90 & 0
& 61.16 & 75.08 & 92.33 & 96.50 & 88.34 & 0 \\

\hline

KNN (Gauss.) 
& \textbf{77.67 ± 0.16} & 85.67 ± 0.13 & 91.71 ± 0.48 & 94.86 ± 0.06 & 88.70 ± 0.36 & 1
& 54.97 & 65.22 & 63.79 & 95.33 & 77.50 & 0
& 56.88 & 66.38 & \textbf{62.05} & \textbf{93.25} & \textbf{76.60} & 3 \\

KNN (Adv.) 
& 77.55 ± 0.18 & \textbf{86.12 ± 0.10} & \textbf{93.24 ± 0.39} & \textbf{95.10 ± 0.03} & \textbf{89.63 ± 0.32} & 4
& 62.64 & 79.67 & \textbf{86.37} & \textbf{97.07} & \textbf{87.02} & 3
& \textbf{56.88} & 66.38 & 62.04 & 93.25 & 76.59 & 1 \\

KNN (Ours) 
& 77.65 ± 0.16 & 85.52 ± 0.13 & 91.23 ± 0.51 & 94.78 ± 0.08 & 88.45 ± 0.38 & 0
& \textbf{63.09} & \textbf{79.87} & 85.99 & 96.91 & 86.90 & 2
& 55.61 & \textbf{65.65} & 63.32 & 94.84 & 77.25 & 1 \\

\hline

PLF (Gauss.) 
& 74.74 ± 0.37 & 80.45 ± 0.19 & 84.30 ± 1.38 & 85.38 ± 0.57 & 82.64 ± 0.80 & 0
& 64.50 & 77.97 & 93.72 & \textbf{95.36} & \textbf{91.07} & 2
& 60.19 & 72.99 & 89.40 & \textbf{92.50} & 85.05 & 1 \\

PLF (Adv.) 
& 76.26 ± 0.30 & 79.61 ± 0.46 & 85.60 ± 0.80 & 78.23 ± 0.31 & 82.00 ± 0.68 & 0
& 71.89 & 78.11 & 91.56 & 90.16 & 89.06 & 0
& 59.95 & 72.15 & 85.79 & 88.98 & 82.27 & 0 \\

PLF (Ours) 
& \textbf{80.31 ± 0.03} & \textbf{85.07 ± 0.17} & \textbf{92.58 ± 0.53} & \textbf{88.70 ± 0.17} & \textbf{88.61 ± 0.32} & 5
& \textbf{73.81} & \textbf{82.34} & \textbf{94.43} & 90.28 & 90.98 & 3
& \textbf{62.82} & \textbf{74.90} & \textbf{90.11} & 90.98 & \textbf{85.31} & 4 \\

\hline

VRA (Gauss.) 
& 73.73 ± 8.46 & 81.57 ± 6.39 & 92.26 ± 1.82 & \textbf{94.31 ± 0.36} & 88.65 ± 3.65 & 1
& 64.44 & 71.15 & \textbf{92.30} & \textbf{96.03} & 87.72 & 2
& 61.33 & 74.04 & \textbf{89.87} & \textbf{93.69} & \textbf{86.17} & 3 \\

VRA (Adv.) 
& \textbf{79.44 ± 0.41} & 85.05 ± 0.49 & 92.83 ± 0.19 & 91.57 ± 2.43 & 89.64 ± 0.93 & 1
& \textbf{72.61} & 79.35 & 89.97 & 86.00 & 87.99 & 1
& 61.42 & 72.84 & 88.83 & 93.37 & 85.43 & 0 \\

VRA (Ours) 
& 79.28 ± 0.04 & \textbf{85.37 ± 0.13} & \textbf{93.37 ± 0.48} & 93.59 ± 0.14 & \textbf{90.73 ± 0.32} & 3
& 70.74 & \textbf{79.85} & 92.16 & 92.82 & \textbf{90.57} & 2
& \textbf{61.84} & \textbf{74.55} & 89.51 & 93.36 & 86.02 & 2 \\

\bottomrule
\end{tabular}
}
\caption{Benchmark results on ImageNet-200 and ImageNet-1k ID datasets using OpenOOD test OOD datasets~\cite{zhang2024openoodv15enhancedbenchmark}. The models are ResNet-18 for ImageNet-200;  ResNet-50 and MobileNet-V2 for ImageNet-1k. We report AUROC (higher is better). Bold indicates the best-performing detector–tuning method for each ID model. Best (\#) denotes the number of test OOD datasets on which a method achieves the highest AUROC. Gauss denotes the Gaussian noise baseline, Adv denotes the adversarial noise baseline, and ours indicates our tuning method.}
\label{tab:imagenet_results}
\end{table*}


\begin{table*}[t]
\centering
\footnotesize
\setlength{\tabcolsep}{4pt}
\resizebox{\textwidth}{!}{%
\begin{tabular}{l|ccc|ccc|ccc|ccc}
\toprule
 & \multicolumn{3}{c|}{\textbf{CIFAR-10} (ResNet-18)} 
 & \multicolumn{3}{c|}{\textbf{CIFAR-100} (ResNet-18)}
 & \multicolumn{3}{c|}{\textbf{ImageNet-200} (ResNet-18)}
 & \multicolumn{3}{c}{\textbf{ImageNet-1K} (ResNet-50)} \\
\cmidrule(lr){2-4} \cmidrule(lr){5-7} \cmidrule(lr){8-10} \cmidrule(lr){11-13}
Method
 & Near-OOD & Far-OOD & ID Acc.
 & Near-OOD & Far-OOD & ID Acc.
 & Near-OOD & Far-OOD & ID Acc.
 & Near-OOD & Far-OOD & ID Acc. \\
\midrule

ReAct & 87.11$\pm$0.61 & 90.42$\pm$1.41 & 95.06$\pm$0.30 
      & \textbf{80.77$\pm$0.05} & 80.39$\pm$0.49 & 77.25$\pm$0.10 
      & 81.87$\pm$0.98 & \textbf{92.31$\pm$0.56} & 86.37$\pm$0.08 
      & 77.38 & \textbf{93.67} & 76.18 \\
\rowcolor{gray!20}
ReAct (Ours) & \textbf{87.53 $\pm$ 0.73} & \textbf{90.97 $\pm$ 1.59} & 95.06$\pm$0.30 
      & 80.52 $\pm$ 0.06 & \textbf{80.76 $\pm$ 0.91} & 77.25$\pm$0.10 
      & \textbf{82.50 $\pm$ 0.07} & 90.89 $\pm$ 0.30 & 86.37$\pm$0.08 
      & \textbf{77.44} & 92.64 & 76.18 \\

\hline

ASH   & 75.27$\pm$1.04 & 78.49$\pm$2.58 & 95.06$\pm$0.30 
      & 78.20$\pm$0.15 & \textbf{80.58$\pm$0.66} & 77.25$\pm$0.10 
      & \textbf{82.38$\pm$0.19} & \textbf{93.90$\pm$0.27} & 86.37$\pm$0.08 
      & 78.17 & \textbf{95.74} & 76.18 \\
\rowcolor{gray!20}
ASH  (ours)  & \textbf{77.99 $\pm$ 2.90} & \textbf{80.71 $\pm$ 4.38} & 95.06$\pm$0.30 
      & \textbf{78.39 $\pm$ 0.23} & 79.33 $\pm$ 1.06 & 77.25$\pm$0.10 
      & 81.95 $\pm$ 0.36 & 92.69 $\pm$ 0.22 & 86.37$\pm$0.08 
      & \textbf{78.22} & 95.26 & 76.18 \\

\hline
KNN   & \textbf{90.64$\pm$0.20} & \textbf{92.96$\pm$0.14} & 95.06$\pm$0.30 
      & \textbf{80.18$\pm$0.15} & 82.40$\pm$0.17 & 77.25$\pm$0.10 
      & 81.57$\pm$0.17 & \textbf{93.16$\pm$0.22} & 86.37$\pm$0.08 
      & 71.10 & \textbf{90.18} & 76.18 \\
\rowcolor{gray!20}
KNN (ours)  & 90.50 $\pm$ 0.19 & 92.77 $\pm$ 0.28 & 95.06$\pm$0.30 
      & 79.91 $\pm$ 0.23 & \textbf{82.44 $\pm$ 0.99} & 77.25$\pm$0.10 
      & \textbf{81.59$\pm$0.15} & 91.49$\pm$0.32 & 86.37$\pm$0.08 
      & \textbf{71.48} & 89.93 & 76.18 \\      
\bottomrule

PLF   & \textbf{89.74$\pm$0.53} & \textbf{92.66$\pm$0.72} & 95.06$\pm$0.30 
      & 80.82$\pm$0.13 & \textbf{80.69$\pm$0.94} & 77.25$\pm$0.10 
      & 82.23$\pm$0.09 & \textbf{93.23$\pm$0.19} & 86.37$\pm$0.08 
      & 76.82 & \textbf{95.27} & 76.18 \\
\rowcolor{gray!20}
PLF (ours)  & 86.75$\pm$0.63 & 90.63$\pm$1.30 & 95.06$\pm$0.30 
      & \textbf{80.93$\pm$0.09} & 80.43$\pm$ 0.74 & 77.25$\pm$0.10 
      & \textbf{82.69$\pm$0.10} & 89.96$\pm$0.34 & 86.37$\pm$0.08 
      & \textbf{78.08} & 91.90& 76.18 \\  
\bottomrule

VRA   
      & \textbf{88.91$\pm$0.68} & \textbf{92.08$\pm$0.67} & 95.06$\pm$0.30 
      & \textbf{80.63$\pm$0.41} & \textbf{81.39$\pm$0.83} & 77.25$\pm$0.10 
      & 82.22$\pm$0.15 & \textbf{93.53$\pm$0.23} & 86.37$\pm$0.08
      & \textbf{77.75} & \textbf{94.89} & 76.18 \\
\rowcolor{gray!20}
VRA (ours)  
      & 87.40$\pm$0.50 & 91.57$\pm$1.24 & 95.06$\pm$0.30 
      & 80.06$\pm$0.15 & 81.12$\pm$1.08& 77.25$\pm$0.10 
      & \textbf{82.32$\pm$0.06} & 92.56$\pm$0.22 & 86.37$\pm$0.08 
      & 75.29 & 91.85 & 76.18 \\
      
\bottomrule

\end{tabular}
}

\caption{Performance comparison of SoA OOD detectors tuned with our method versus given test OOD data used in the Open OOD benchmark. We compare the performance of OOD detectors tuned with our approach to the detectors tuned with given OOD data, specifically selected in OpenOOD benchmarks to have favorable properties for tuning. Detectors tuned by our method using only in-distribution data sometimes out-performs the same detectors tuned with given OOD data, and in the cases it doesn't, our method is generally comparable in performance.  OOD detection results for ReAct, ASH, KNN, VRA and PLF across CIFAR and ImageNet benchmarks in OpenOOD are shown. In OpenOOD, all the detectors are tuned using validation subsets from TinyImageNet for CIFAR 10/100 and OpenImage-O for ImageNet 200/1k. “Ours” refers to our proposed tuning method for each detector. Near-OOD and Far-OOD columns correspond to average AUROC of the detection on the identified near and far OOD datasets of each benchmark as reported in OpenOOD. }
\label{tab:table_val_set_tuning_comparison}
\end{table*}

\subsection{Datasets and Settings}
We evaluate our method using protocols and data from the OpenOOD benchmark \cite{zhang2023openood}. We experiment with ResNet-18\cite{he2015deepresiduallearningimage},  ResNet-50\cite{Resnet50}, MobileNet-v2~\cite{mobilenetv2} architectures, and the following in distribution datasets.

\textbf{CIFAR-10}
We conduct experiments on CIFAR-10 \allowbreak\cite{krizhevsky2009learning}, which contains $n=10$ categories. 
To construct simulated OOD categories, we randomly select 
$M_{\text{ood,sim}} \in \{1,2,3,4,5\}$ categories as simulated OOD classes and treat the remaining 
$M_{\text{id,sim}} = n - M_{\text{ood,sim}}$ categories as simu-
lated ID categories.
For each value of $M_{\text{ood,sim}}$, we repeat the random category selection using ten random 
seeds $\{0,1 \hdots 9\}$, resulting in $5 \times 10 = 50$ simulated ID/OOD splits. Using $\mathcal{D}_{\text{id,sim},i}^M$ and $\mathcal{D}_{\text{ood,sim},i}^M$, together with additional samples drawn from $\mathcal{D}_{\text{ood,sim},i}^t$ when necessary, we construct a balanced ID/OOD training dataset for the OOD detector.

\textbf{CIFAR-100} \cite{krizhevsky2009learning} contains $n=100$ categories, and we follow similar experimental setup as CIFAR-10, with the only difference being the number of simulated OOD categories. Here we select eight
$M_{\text{ood,sim}} \in \{5, 10, 15, \hdots 40\}$ categories as possible number of simulated OOD classes. For each value of $M_{\text{ood,sim}}$, we repeat the random category selection using ten random 
seeds $\{0,1 \hdots 9\}$, following the same procedure as CIFAR-10.

\textbf{ImageNet-1K.} ~\cite{russakovsky2015imagenet}  contains $n=1000$ categories. 
We select \linebreak[3] $M_{\text{ood,sim}} \in \{50,100,\allowbreak150,\allowbreak\ldots,\allowbreak300\}$ categories as simulated OOD classes.
For each setting, we generate splits using five random seeds $\{0,1,\ldots,4\}$ and follow the same balanced dataset construction procedure as in CIFAR-10.

\textbf{ImageNet-200} ~\cite{zhang2024openoodv15enhancedbenchmark} follows a similar experimental setup. 
We select 
$M_{\text{ood,sim}} \in \{10,20,30,40\}$ categories as simulated OOD classes and generate splits using three random seeds $\{0,1,2\}$.
The remaining categories are treated as simulated ID classes, and balanced training datasets are constructed following the same procedure as above. 

\textbf{Gaussian Noise} We follow~\citep{kirchheim2022pytorch} to generate Gaussian noise images for CIFAR and ImageNet benchmarks. We sample the noise images as $\mathcal{N (\mu, \sigma)}$ at each pixel location, and thereafter clip each pixel to be in the range [0, 255]. For CIFAR benchmarks, the noise images generated are 32 $\times$ 32 $\times$ 3, and 224 $\times$ 224 $\times$ 3 for the ImageNet benchmarks. We fix $\mu = 128$, and consider three levels for the hyperparameter $\sigma$ : 32, 64 and 128.

\textbf{Adversarial Perturbations} Given an input image $x$ from CIFAR and ImageNet datasets and its ground-truth label $y$, adversarial examples are generated using the Fast Gradient Sign Method (FGSM) \cite{goodfellow2015explaining}:
\begin{equation}
x_{\mathrm{adv}} = x + \epsilon \cdot \mathrm{sign}\big(\nabla_x \mathcal{L}(x, y)\big),
\end{equation}
where $\mathcal{L}$ denotes the Binary Cross-Entropy (BCE) loss and $\epsilon$ controls the perturbation magnitude.
We consider three levels for the hyperparameter $\epsilon$: $0.005$, $0.01$, and $0.1$. After applying the perturbation, the adversarial images are clipped to remain within the valid pixel range.

We choose to optimize AUROC as our loss function, as is standard for OOD tuning in the literature.

\subsection{OOD Detectors Trained} \label{Sec: OOD_detectors_tuned} We evaluate our tuning framework on a representative set of post-hoc OOD detectors: ReAct~\cite{sun2021reactoutofdistributiondetectionrectified}, ASH~\cite{djurisic2022ash}, KNN~\cite{sun2022knn}, VRA~\cite{xu2023vravariationalrectifiedactivation}, and PLF~\cite{mondal2025variationalinformationtheoreticapproach}. These methods are all leading state-of-the-art post-hoc inference methods~\cite{zhang2024openoodv15enhancedbenchmark} applied to a particular layer of a trained neural network, and operate by transforming or re-weighting intermediate feature representations before computing an OOD score. Of these, ASH, ReAct and KNN have one tuning parameter each. ReAct applies element-wise activation clipping and is controlled by a single threshold parameter, while ASH performs sparsification based on a percentile parameter that determines how many activations are suppressed. KNN estimates OOD scores using distances in feature space and requires tuning the number of nearest neighbors used for scoring. On the other hand, VRA and PLF are higher parameter piecewise linear family of detectors. VRA~\cite{xu2023vravariationalrectifiedactivation} is parameterized by three values that control the suppression of low activations, the amplification of mid-range activations, and the clipping of high activations. PLF is characterized by seven parameters which denote a more general piecewise linear family of detectors~\cite{mondal2025variationalinformationtheoreticapproach}. More details of the tuning parameters and their ranges used for optimization are provided in the Appendix~\ref{app:HyperOpt}. 

\subsection{Results}

We report the results of our tuning method on CIFAR 10/100  benchmarks (Table~\ref{tab:cifar_results}) and the Imagenet 200/1k benchmarks (Table~\ref{tab:imagenet_results}) using the OOD detectors mentioned in Section~\ref{Sec: OOD_detectors_tuned}. We compare our tuning strategy against Gaussian-noise and adversarial-perturbation tuning, which serve as baseline approaches.

We notice that our approach generally performs better than the other tuning methods on the higher parameter OOD detectors - VRA and PLF. 
For the CIFAR benchmarks our approach achieves the best AUROC on 10 datasets each for VRA and PLF, substantially improving over Gaussian and adversarial tuning. For PLF, our method achieves the best AUROC on all 5 OOD datasets for ImageNet-200 (ResNet-18) and ImageNet-1K (ResNet-50), and on 4 out of 5 datasets for ImageNet-1K (MobileNet-V2). These results indicate that our tuning strategy can be particularly effective for detectors with more complex parameterizations, whereby the detectors may overfit to Gaussian noise or adversarial perturbations as OOD tuning data, and can potentially fail to generalize under test settings. 

For other lower complexity shaping approaches such as ReAct and ASH, our approach remains competitive and often achieves best or second-best performance across datasets. In particular, on CIFAR benchmarks, our method achieves the highest number of wins for ReAct compared to its Gaussian and adversarial counterparts, and remains competitive for ASH, often matching or improving upon adversarial tuning across datasets. The observations are similar for KNN, which relies primarily on fixed feature geometry and has limited tunable capacity. On CIFAR-10, our method consistently ranks best with KNN, while on CIFAR-100 and ImageNet benchmarks the results are mixed, with our approach frequently ranking first or second and often outperforming Gaussian noise tuning. Overall, these results suggest that our OOD detector tuning approach yields the largest and most consistent improvements for higher-capacity detectors such as PLF and VRA, while providing more modest and less uniform gains for lower-capacity methods.

We also compare our results with those reported in OpenOOD benchmark~\cite{zhang2024openoodv15enhancedbenchmark} in Table~\ref{tab:table_val_set_tuning_comparison}. OpenOOD specifies a given OOD tuning set from TinyImageNet for CIFAR 10/100, and OpenImage-O for ImageNet 200/1k on which the hyperparameters of the OOD detectors are tuned. These tuning sets are specifically chosen for favorable properties as tuning sets. Even though our tuning approach only uses in-distribution data, the results in  Table~\ref{tab:table_val_set_tuning_comparison} show that OOD detectors tuned with our method sometimes out performs the detectors tuned with given OOD data, and in cases that it doesn't the results are generally comparable. This suggests our method can be used when no OOD tuning data is available, without an appreciable performance loss.


\subsection{Ablation on number of simulated OOD categories}

We present an ablation on the number of simulated OOD categories for our experiments on ImageNet-200 and ImageNet-1k, in Tables~\ref{tab:imagenet200_resnet18_holdout} and~\ref{tab:imagenet1k_holdout}, respectively. For most methods, AUROC varies only slightly across different holdout sizes. Minor fluctuations are method-dependent, but overall the results demonstrate robustness of our tuning approach over a range of simulated OOD categories for both the benchmarks. The observations are similar for the CIFAR benchmarks, and are therefore omitted for brevity.

\begin{table}[h]
\centering
\small
\resizebox{1\linewidth}{!}{
\begin{tabular}{lcccccccc}
\toprule
 & \multicolumn{7}{c}{ImageNet-200 (ResNet-18)} \\
\cmidrule(lr){2-8}
Method & Holdout & SSB-hard & NINCO & iNaturalist & Textures & OpenImage-O & AVG \\
\midrule
ReAct & 10 & 79.83 $\pm$ 0.02 & 85.17 $\pm$ 0.12 & 92.56 $\pm$ 0.50 & 90.87 $\pm$ 0.16 & 89.25 $\pm$ 0.25 & 87.536 \\
      & 20 & 79.83 $\pm$ 0.02 & 85.17 $\pm$ 0.11 & 92.55 $\pm$ 0.50 & 90.79 $\pm$ 0.16 & 89.23 $\pm$ 0.26 & 87.514 \\
      & 30 & 85.17 $\pm$ 0.11 & 82.50 $\pm$ 0.05 & 90.79 $\pm$ 0.16 & 89.23 $\pm$ 0.26 & 90.86 $\pm$ 0.21 & 87.710 \\
      & 40 & 79.83 $\pm$ 0.02 & 85.17 $\pm$ 0.11 & 92.55 $\pm$ 0.50 & 90.79 $\pm$ 0.16 & 89.23 $\pm$ 0.26 & 87.514 \\
\midrule
ASH   & 10 & 79.38 $\pm$ 0.29 & 84.51 $\pm$ 0.43 & 94.52 $\pm$ 0.41 & 93.14 $\pm$ 0.17 & 90.42 $\pm$ 0.09 & 88.394 \\
      & 20 & 79.74 $\pm$ 0.25 & 85.18 $\pm$ 0.51 & 94.91 $\pm$ 0.46 & 93.86 $\pm$ 0.10 & 91.21 $\pm$ 0.14 & 88.980 \\
      & 30 & 79.86 $\pm$ 0.21 & 85.48 $\pm$ 0.58 & 95.04 $\pm$ 0.46 & 94.33 $\pm$ 0.08 & 91.59 $\pm$ 0.18 & 89.260 \\
      & 40 & 79.86 $\pm$ 0.21 & 85.48 $\pm$ 0.58 & 95.04 $\pm$ 0.46 & 94.33 $\pm$ 0.08 & 91.59 $\pm$ 0.18 & 89.260 \\
\midrule
KNN   & 10 & 77.65 $\pm$ 0.16 & 85.52 $\pm$ 0.13 & 91.23 $\pm$ 0.51 & 94.78 $\pm$ 0.08 & 88.45 $\pm$ 0.38 & 87.526 \\
      & 20 & 77.65 $\pm$ 0.16 & 85.52 $\pm$ 0.13 & 91.23 $\pm$ 0.51 & 94.78 $\pm$ 0.08 & 88.45 $\pm$ 0.38 & 87.526 \\
      & 30 & 77.42 $\pm$ 0.20 & 86.17 $\pm$ 0.10 & 93.58 $\pm$ 0.37 & 95.17 $\pm$ 0.02 & 89.88 $\pm$ 0.31 & 88.444 \\
      & 40 & 77.54 $\pm$ 0.19 & 86.13 $\pm$ 0.10 & 93.28 $\pm$ 0.38 & 95.11 $\pm$ 0.03 & 89.66 $\pm$ 0.32 & 88.344 \\
\midrule
PLF   & 10 & 80.31 $\pm$ 0.03 & 85.07 $\pm$ 0.17 & 92.58 $\pm$ 0.53 & 88.70 $\pm$ 0.17 & 88.61 $\pm$ 0.32 & 87.054 \\
      & 20 & 79.99 $\pm$ 0.02 & 85.10 $\pm$ 0.11 & 92.46 $\pm$ 0.48 & 89.54 $\pm$ 0.17 & 88.89 $\pm$ 0.24 & 87.196 \\
      & 30 & 79.85 $\pm$ 0.10 & 85.47 $\pm$ 0.14 & 93.37 $\pm$ 0.38 & 91.29 $\pm$ 0.17 & 89.89 $\pm$ 0.24 & 87.974 \\
      & 40 & 80.05 $\pm$ 0.01 & 85.57 $\pm$ 0.08 & 93.02 $\pm$ 0.45 & 91.01 $\pm$ 0.16 & 89.69 $\pm$ 0.18 & 87.868 \\
\midrule
VRA   & 10 & 79.25 $\pm$ 0.06 & 85.03 $\pm$ 0.16 & 92.84 $\pm$ 0.53 & 92.56 $\pm$ 0.15 & 90.06 $\pm$ 0.36 & 87.948 \\
      & 20 & 79.26 $\pm$ 0.04 & 85.40 $\pm$ 0.13 & 93.40 $\pm$ 0.48 & 93.48 $\pm$ 0.14 & 90.67 $\pm$ 0.33 & 88.442 \\
      & 30 & 79.22 $\pm$ 0.05 & 85.30 $\pm$ 0.14 & 93.27 $\pm$ 0.49 & 93.26 $\pm$ 0.13 & 90.51 $\pm$ 0.33 & 88.312 \\
      & 40 & 79.28 $\pm$ 0.04 & 85.37 $\pm$ 0.13 & 93.37 $\pm$ 0.48 & 93.59 $\pm$ 0.14 & 90.73 $\pm$ 0.32 & 88.468 \\
\bottomrule
\
\end{tabular}}
\caption{AUROC on ImageNet-200 (ResNet-18) under Different Holdout-Class Settings}
\label{tab:imagenet200_resnet18_holdout}
\end{table}

\begin{table*}[t]
\centering
\small
\resizebox{\textwidth}{!}{
\begin{tabular}{l c cccccc cccccc}
\toprule
& \multicolumn{13}{c}{ImageNet-1K} \\
\cmidrule(lr){3-14}
& \multicolumn{7}{c|}{ResNet-50}
& \multicolumn{6}{c}{MobileNet-V2} \\
\cmidrule(lr){3-8}
\cmidrule(lr){9-14}
Method & Holdout
& SSB-hard & NINCO & iNaturalist & Textures & OpenImage-O & AVG
& SSB-hard & NINCO & iNaturalist & Textures & OpenImage-O & AVG \\
\midrule
ReAct & 50  & 73.15 & 81.76 & 95.05 & 91.87 & 91.44 & 86.66 & 61.62 & 73.49 & 91.88 & 93.49 & 86.38 & 81.37 \\
      & 100 & 72.48 & 81.39 & 94.23 & 91.14 & 90.92 & 86.14 & 61.54 & 73.41 & 91.53 & 93.21 & 86.12 & 81.16 \\
      & 150 & 73.18 & 81.69 & 94.88 & 91.72 & 91.33 & 86.56 & 61.63 & 73.50 & 91.90 & 93.50 & 86.39 & 81.38 \\
      & 200 & 73.61 & 81.65 & 94.78 & 91.63 & 91.27 & 86.50 & 61.69 & 73.51 & 92.16 & 93.69 & 86.56 & 81.52 \\
      & 250 & 73.00 & 81.55 & 94.56 & 91.43 & 91.13 & 86.36 & 61.69 & 73.51 & 92.18 & 93.71 & 86.58 & 81.54 \\
      & 300 & 73.27 & 81.76 & 95.04 & 91.87 & 91.43 & 86.66 & 61.15 & 75.01 & 92.21 & 96.46 & 88.27 & 82.62 \\
\midrule
ASH   & 50  & 73.15 & 83.30 & 96.73 & 96.48 & 93.04 & 88.54 & 66.99 & 78.17 & 92.09 & 95.73 & 90.41 & 84.68 \\
      & 100 & 73.20 & 83.26 & 96.69 & 96.41 & 93.00 & 88.51 & 61.04 & 74.59 & 91.58 & 96.26 & 87.88 & 82.27 \\
      & 150 & 73.05 & 83.37 & 96.83 & 96.63 & 93.17 & 88.61 & 61.19 & 75.17 & 92.53 & 96.56 & 88.46 & 82.78 \\
      & 200 & 73.11 & 83.31 & 96.76 & 96.54 & 93.09 & 88.56 & 61.16 & 75.08 & 92.33 & 96.50 & 88.34 & 82.68 \\
      & 250 & 73.27 & 83.17 & 96.60 & 96.29 & 92.90 & 88.45 & 61.15 & 75.05 & 92.30 & 96.49 & 88.33 & 82.67 \\
      & 300 & 73.05 & 83.34 & 96.81 & 96.61 & 93.16 & 88.60 & 61.73 & 73.50 & 92.33 & 93.81 & 86.66 & 81.61 \\
\midrule
KNN   & 50  & 61.86 & 79.27 & 86.78 & 97.27 & 87.10 & 82.46 & 56.68 & 66.28 & 62.30 & 93.57 & 76.73 & 71.11 \\
      & 100 & 61.74 & 79.20 & 86.82 & 97.30 & 87.10 & 82.43 & 55.98 & 65.32 & 62.93 & 95.15 & 77.49 & 71.38 \\
      & 150 & 63.09 & 79.87 & 85.99 & 96.91 & 86.90 & 82.55 & 56.89 & 66.39 & 62.05 & 93.25 & 76.60 & 71.03 \\
      & 200 & 63.09 & 79.87 & 85.99 & 96.91 & 86.90 & 82.55 & 55.20 & 65.39 & 63.63 & 95.17 & 77.41 & 71.36 \\
      & 250 & 63.09 & 79.87 & 85.99 & 96.91 & 86.90 & 82.55 & 56.68 & 66.28 & 62.30 & 93.57 & 76.73 & 71.11 \\
      & 300 & 63.09 & 79.87 & 85.99 & 96.91 & 86.90 & 82.55 & 55.61 & 65.65 & 63.32 & 94.84 & 77.25 & 71.33 \\
\midrule
PLF   & 50  & 73.63 & 82.90 & 95.40 & 92.45 & 91.98 & 87.27 & 60.68 & 73.04 & 87.35 & 89.51 & 83.25 & 78.77 \\
      & 100 & 73.98 & 82.65 & 94.87 & 91.19 & 91.45 & 86.83 & 62.88 & 75.13 & 90.75 & 91.17 & 85.70 & 81.12 \\
      & 150 & 73.81 & 82.34 & 94.43 & 90.28 & 90.98 & 86.37 & 62.79 & 73.74 & 89.50 & 88.80 & 84.01 & 79.77 \\
      & 200 & 73.80 & 81.47 & 93.03 & 87.55 & 89.59 & 85.09 & 62.69 & 74.46 & 90.27 & 90.48 & 85.12 & 80.61 \\
      & 250 & 73.85 & 82.16 & 94.10 & 89.71 & 90.71 & 86.11 & 63.27 & 74.29 & 90.63 & 87.98 & 84.06 & 80.05 \\
      & 300 & 73.80 & 81.47 & 93.03 & 87.55 & 89.59 & 85.09 & 62.82 & 74.90 & 90.11 & 90.98 & 85.31 & 80.82 \\
\midrule
VRA   & 50  & 70.33 & 80.28 & 92.98 & 94.00 & 91.23 & 85.76 & 61.84 & 74.79 & 89.92 & 93.73 & 86.41 & 81.34 \\
      & 100 & 70.34 & 80.22 & 92.88 & 93.87 & 91.15 & 85.69 & 61.88 & 74.00 & 88.99 & 92.35 & 85.24 & 80.49 \\
      & 150 & 70.74 & 79.85 & 92.16 & 92.82 & 90.57 & 85.23 & 61.96 & 73.42 & 88.51 & 91.06 & 84.35 & 79.86 \\
      & 200 & 70.70 & 79.85 & 92.18 & 92.86 & 90.58 & 85.24 & 61.84 & 74.55 & 89.51 & 93.36 & 86.02 & 81.05 \\
      & 250 & 71.05 & 79.91 & 92.14 & 92.62 & 90.53 & 85.25 & 61.97 & 73.43 & 88.53 & 91.08 & 84.37 & 79.87 \\
      & 300 & 70.94 & 79.87 & 92.12 & 92.65 & 90.52 & 85.22 & 61.96 & 73.38 & 88.44 & 91.00 & 84.28 & 79.81 \\
\bottomrule
\end{tabular}}
\caption{AUROC on ImageNet-1K under different holdout-class settings.}
\label{tab:imagenet1k_holdout}
\end{table*}

\section{Limitations}
Our method involves training $N$ variants of the task neural network; while manageable with standard computational hardware (e.g., workstations equipped with Nvidia RTX 3090 GPUs) for the classification task, this can be computationally expensive for tasks that involve dense predictions, e.g., segmentation, as well as large language models. Furthermore, we tested convolutional neural network architectures, where each experiment took roughly 30 minutes on our workstation. We note that these architectures are comparably less expensive than transformer-based models. Future work will aim to address this limitation, and extend our formulations for other tasks.


\section{Conclusion}
We showed through extensive benchmarking that OOD detectors, especially larger parameter detectors, can exhibit large variance across OOD tuning sets, in particular those chosen in existing benchmarks. This makes the OOD detectors sensitive to the predefined tuning set. Moreover, obtaining such a tuning set in practice may be difficult. We formulated the new problem of tuning OOD parameters without a given (real) OOD tuning data. Compared to strong baselines that OOD practioners might use to tune OOD detectors without collected tuning data, our new method out-performs on higher parameter families of detectors and are comparable in lower parameter detectors. This was demonstrated on numerous datasets and OOD detectors. Our method thus provides a path to deploying reliable OOD detectors in practical applications. Furthermore, this work suggests that further research in tuning OOD detectors without given OOD data is needed.

\section*{Impact Statement}

This paper presents work whose goal is to advance the field of Machine Learning. There are many potential societal consequences 
of our work, none which we feel must be specifically highlighted here.


\bibliographystyle{icml2026}

\newpage
\appendix
\onecolumn
\section{Detector Definitions and Parameter Selection}
\label{app:HyperOpt}
\paragraph{ReAct.}
ReAct \cite{sun2021reactoutofdistributiondetectionrectified} clips penultimate-layer feature activations element-wise:
\[
\mathrm{ReAct}(z)_i = \min(z_i, \tau).
\]
The single parameter $\tau$ is optimized by Bayesian optimization. The search interval is defined from the range of activations from the penultimate layer obtained with the ID data, similar to the setup of~\cite{sun2021reactoutofdistributiondetectionrectified}. 

\paragraph{ASH.}
ASH \cite{djurisic2022ash} sparsifies intermediate activations by retaining the top-$k$ elements and suppressing the rest, where $k$ is determined by a percentile parameter $p$. The retained activations are set to a constant proportional to the activation energy. The only tunable parameter is $p$, which is optimized by Bayesian optimization over a valid percentile range on simulated ID/OOD splits. We use the ASH-B variant of ASH in our benchmarks.

\paragraph{VRA.}
VRA+ \cite{xu2023vravariationalrectifiedactivation} applies a piecewise shaping function:
\[
\mathrm{VRA}(z) =
\begin{cases}
0, & z < \alpha,\\
z + \gamma, & \alpha \le z \le \beta,\\
\beta, & z > \beta .
\end{cases}
\]
Rather than optimizing $\alpha$ and $\beta$ directly, we parameterize them by quantiles of the empirical ID feature distribution:
\[
\alpha = F^{-1}_{\text{ID}}(\eta_\alpha), \quad 
\beta = F^{-1}_{\text{ID}}(\eta_\beta).
\]
The optimization variables are $(\eta_\alpha, u, \gamma)$ with
\[
\eta_\alpha \in [0.1, 0.8], \quad u \in [0,1], \quad \gamma \in [0,5],
\]
and $\eta_\beta = \eta_\alpha + \delta$, where
\[
\delta = \delta_{\min} + u(\delta_{\max}-\delta_{\min}), \quad
\delta_{\min}=0.10, \quad \delta_{\max}=0.99-\eta_\alpha,
\]
ensuring $\eta_\beta > \eta_\alpha$. Bayesian optimization selects $(\eta_\alpha,\eta_\beta,\gamma)$ on simulated ID/OOD splits.

\paragraph{PLF.}
PLF applies a seven-parameter piecewise linear shaping function to features~\cite{mondal2025variationalinformationtheoreticapproach}, where $f(\cdot)$ consists of three linear segments defined by breakpoints $x_1,x_2$ and slopes $m_1,m_2$ with vertical offsets $(y_{\text{start}}, y_{\text{end}}, y_1)$. The breakpoints are defined by quantiles of the absolute ID features:
\[
x_1 = F^{-1}_{|Z_{\text{ID}}|}(q_1), \quad
x_2 = F^{-1}_{|Z_{\text{ID}}|}(q_2),
\]
with $q_2 = q_1 + \delta$. The optimization variables are
\[
(y_{\text{start}}, y_{\text{end}}, \Delta y, q_1, u, m_1, m_2),
\]
with
\[
y_{\text{start}} \in [-5,0], \quad y_{\text{end}} \in [0,5], \quad \Delta y \in [0,5],
\]
\[
q_1 \in [0.1,0.8], \quad u \in [0,1], \quad m_1 \in [0,5], \quad m_2 \in [-5,5],
\]
and $y_1 = y_{\text{end}}+\Delta y$. The upper quantile is defined as
\[
\delta = \delta_{\min} + u(\delta_{\max}-\delta_{\min}), \quad
\delta_{\min}=0.10, \quad \delta_{\max}=0.99-q_1,
\]
so that $q_2 = q_1 + \delta$ and $q_2>q_1$. Parameters are selected by Bayesian optimization on simulated ID/OOD splits.

\paragraph{KNN.}
KNN \cite{sun2022knn} computes OOD scores using the negative distance to the $K$-th nearest neighbor in normalized feature space. The only tunable parameter is $K$. We search over integer $K \in [1, K_{\max}]$ using Bayesian optimization, where distances are precomputed from a feature index built on ID training data. We choose $K_{max} = 500$

\paragraph{Optimization Protocol.}
For all detectors, parameters are selected using Bayesian optimization with a Gaussian process surrogate on simulated ID/OOD splits. The objective is to maximize AUROC computed from energy-based scores after applying the corresponding feature transformation.

\end{document}